# Does Syntactic Knowledge help English-Hindi SMT ?

**Taraka Rama,Karthik Gali,Avinesh PVS**
Language Technologies Research Centre,
{taraka,karthikg,avinesh}@students.iiit.ac.in

## Abstract

In this paper we explore various parameter settings of the state-of-art Statistical Machine Translation system to improve the quality of the translation for a 'distant' language pair like English-Hindi. We proposed new techniques for efficient reordering. A slight improvement over the baseline is reported using these techniques. We also show that a simple pre-processing step can improve the quality of the translation significantly.

## 1 Introduction

Using Statistical Machine Translation techniques for translating between European languages has been quite successful in the last few years. MOSES(Koehn et al., 2007), the SMT tool kit is used very frequently in the machine translation experiments and also for deploying real time systems. A glimpse of papers being published at any major NLP conference shows that the techniques have been widely popular not only between linguistically related European languages but also between English-Arabic and English-Chinese , which are linguistically distant pairs. There has been some work done(Venkatapathy and Bangalore, ) on the English-Hindi Machine translation but there is lot more to be explored. In this paper we propose new reordering approaches for improving the translation quality. The translation quality is evaluated using the widely popular Bleu metric(Papineni et al., 2001). We compare ourselves with the baseline score and report the improvements. The paper has a brief outline as follows. In Section 2 we identify the problem areas, Section 3 elaborates on the approach we adopted, Section 4 briefs about the data sets, Section 5 talks about initial experiments with different Moses settings. Section 6 gives a picture of the results and the various results performed and in Section 7 we make observations based on the experiments' results and Section 8 discusses the future directions to be taken and the new subareas to be explored.

## 2 Scope for Improvement

We compare the Bleu score of English-hindi with English-German, English-French in Table 1. The Bleu score for the English-French and English-German language pairs were obtained from the Statistical Machine Translation website[1]. For the English-Hindi pair, the Bleu score is reported as given in the Shared Task website[2].

The differences in the sizes of the training sets have to be kept in mind, when reporting the Bleu score of the translated language pairs. The huge difference in the Bleu scores of English-French and English-Hindi is due to the sizes of the training corpora. When normalising the sizes of the training corpora, we can observe that the real difference in the Bleu scores. At this point we only have a single reference translation. Ideally, the system has to be compared against the multiple refernce data sets and then Bleu score should be an average of all the Bleu scores. One obvious reason for the lower Bleu score is the linguistic distance between the two languages. The underlying structural differences between the two languages manifest themselves as a relatively low score. The structural differences can be broken into smaller subproblems and then can be addressed separately. We have identified the following problem areas where there is a huge scope for improvization. The most significant difference is in the word order (chunk order) of the two languages.

---

[1] http://www.statmt.org/matrix/
[2] http://ltrc.iiit.ac.in/nlptools2008/resources2.php

| **Language-Pair** | Bleu Score |
|---|---|
| **English-French** | 31.36 |
| **English-German** | 25.36 |
| **English-Hindi** | 17.70 |

Table 1: Comparision of Bleu Scores for Various language Pairs

So the problem corresponding to this difference is the issue of reordering. This forms a major weighting factor for the low Bleu score. Infact the performance of the Moses, when it comes to word to word translation is quite good. When we look into the Unigram Bleu score the score is as high as 60 which corroborates this claim. But there is a problem of unknown words in the output in the target language. So the major thrust in the course of our experiments was to tackle this issue of reordering. Nonethless, the problem of unknown word translation is also examined.

## 3 Our Approach

Our hypothesis is that instead of trying to reorder the target language ouput using POS tags or huge language models or chunk language models, we instead go for the rearrangement of the words in the source language itself. The quality can also be improved by using richer and huge language models. This in itself, is a costly procedure. Especially, for a language like English which has syntactic parsers of high quality, it is always desirable to tap these existing resources. Our modelling goes as follows. The core idea is that learning the reordering relations from the parse of the source language sentence would fetch better results. The source sentence is parsed and hence reordered by learning the source-target reordering relations using POS tags on both the sides. This scheme would ensure that the transfer rules learnt are more generic and hence the intution that they would improve the performance of the system.

The second hypothesis is concerning the presence of the unknown words in the target language output. Two possible ways are factored models as given in(Koehn and Hoang, 2007) and using a bilingual dictionary to translate the lemmas and if the unknown word is a named entity then transliterate it. Factored models are designed to overcome the limitations of the phrase-based translations. The factored models make use of lemmas, POS tags and morphology to generate the surface forms while translating. Recently, factored models are becoming famous and hence we opted out to test the factored models using POS taggers and morph analysers, which are available for both English and Hindi side. We toyed with the distortion limit and we observed that when we allow unlimited translations the Bleu score improved. But the improvement was not so much. Mention the Bleu score. Here again we found no significant improvement over the old models. The hypothesis is falsified if the Bleu score shows no improvement.

## 4 Data Sets and Baseline

We initially explored various parameters of the MOSES to see how the quality improves. The training data for learning word alignments was conducted on 7000 sentence parallel corpora provided in the Shared Task(taken from EILMT corpus). We also went on to check how the accuracy increases by increasing the size of the monolingual corpus on which language model is trained. For the baseline the following parameters were adopted. The distortion limit was kept at 6 and the lexicalised reordering models were trained by using msd-bidirectional-fe. The distortion limit shows how often the phrases get reordered. The MOSES has both a distance based reordering model and a lexicalised reordering model. Lexicalised reordering model gives the orientation of a phrase. The distortion limit is simply the absolute of the difference of the last word of the previously translated English phrase and the position of the first word in the current phrase. The language model was trained on the hindi side of the training data set which has 7000 sentences. We have taken a language model of order 5 which was trained using SRILM toolkit(Stolcke, 2002). The lexicalised reordering models give the orientation of a particular phrase while training. Its either that when the phrase is being translated if there a alignment point is found to the left of the current phrase then the orientaton is mono otherwise if the next alignment point is found to the right then the orientation is swap. The models are trained bidirectionally and on both sides. The alignment heuristics used were align-grow-diag-final. Further details can be obtained from (Och and Ney, 2003).

## 5 Initial Experiments

We observed that decreasing the maximum phrase length does not improve the Bleu score. We proceeded to test if the lexicalised reordering models really contribute to the Bleu score. We observed that removing the lexicalised ordering decreases the Bleu score. We also toyed with the distortion limit parameter. We observed that allowing unlimited distortions improves the Bleu Score. Hence the distortion limit was always kept at -1 for allowing unlimited reorderings. We have conducted the following set of experiments. Note that all the experiments which we have conducted can be categorised as pre-processing steps. The motive was to examine if we can tap all the available rich resources to do some pre-processing to achieve improvement in the Bleu score. Apart from the experiments with parser and bilingual dictionary we also checked whether any other pre-processing steps could improve the Bleu score.

At this stage we have gone for another factored model experiment where, instead of guessing the POS tag through the morph information we go for direct lemma to lemma translation and morph-morph translaton. Then the morph information and the lemma are used to guess the POS tag and then the POS tags, morph and the lemma are used to generate the surface forms. Although a significant improvement has been made still the Bleu score was far away from the baseline. We used the morphological analyzers developed as a part of SHAKTI project(Bharati et al., 2003) to extract the morph information and lemma and POS tags. We also experimented by changing the maximum phrase length to 3 and then 4 as given in(Koehn et al., 2003). But there was no improvement made in the translation. We also experimented with the Minimum Bayes Risk Decoder(Kumar et al., 2004) whose 'Loss Function' is based on the Bleu scoring function itself.

## 6 Results

We tried out the following set of experiments. The morph information guesses the POS tag, the lemma is translated, lemma is also used to guess the POS tag of the target side. In the end the guessed POS tag and the translated lemma are used to generate the surface form. The results were not good enough and did not even reach the base line. Next we tried outanother experiment using another configuration of the factored models to improve the translation. All this is in an attempt to improve the translation of the verbs. The english verbs either occur as a single form or with the auxillary verb preceding the main verb. In Hindi the auxillary verbs follow the main verb in the form of a copula at the end. For example the "is" verb is word aligned to "heM" or "hai" or "thA". We have observed that the preposition reordering is fine. In all this experiments we assumed that the word to word alignemnts are correct. This task has to be addressed in a separate paper.

### 6.1 Learning reorderings from Parser

The objective was to incorporate as much as syntactic information possible to affect the reordering. So we tried to learn the rules of reordering which govern the reorderings from the parallel corpus. We wanted to tap the resources available for english as much as possible. Here we used Libin's dependency parser(Shen, 2006) to learn the structure of the english sentence. Then the target POS order is taken to learn the word order rules. These rules were then applied on the english training data and then the training is done as usual. The following type of rules have been learnt.

```
VB  ---  VB
(6)      (6)
 IN   NN-1  ---  NN1    IN
 (4)  (5)       (4)    (5)
IN~1_NN&_VB~2 ==> NN&_IN~1_VB~2
```

We achieved the baseline's score at this point. At this point we used the additional hindi corpora of 7,000 hindi sentences for training the language model and could get a improvement of 0.65 over the base line. When the dictionary was incorporated in the reordering model, there was a slight improvement of 0.05 in the Bleu score.

### 6.2 Dictionary Experiments

The English-Hindi bilingual dictionary was used to translate the unknown words given by the system. The root of the english words was simply replaced by the corresponding hindi root as a part of incorporating the dictionary. When the bilin-gual dictionary was applied on the unknown words on the tuned output there was no improvement in the Bleu score over the baseline model. However an increase in the Bleu score was observed when the dictionary was used to replace the unknown words in the output of the reordering phase. Table 2 shows the results of this experiment. As expected when the language model was trained using additional corpora there was an increase in the Bleu score.

| System | Bleu Score |
|---|---|
| **Baseline** | 17.70 |
| **Parser** | 17.70 |
| **Dictionary** | 17.75 |
| **Language Model** | 18.35 |

Table 2: Results of Dictionary and Parser Steps

| System | Bleu Score |
|---|---|
| **Baseline** | 17.70 |
| **Preprocessing** | 18.09 |
| **After Tuning** | 19.98 |

Table 3: Results of Preprocessing steps

### 6.3 Preprocessing Experiments

The initial experiment was to remove the end of sentence markers. We only removed the end of sentence markers of the declarative sentences. No other type of end of sentence markers were removed. This was done on the training, development as well as the test set. The markers can be added at the end after the translation is complete. This experiment improved the Bleu score by 0.39 on test set. This experiment was conducted without any tuning.

### 6.4 Tuning

Then we tuned our trained model parameters on the development set and then tested the model on the previously used test set. The tuning was done using the Minimum Error Rate Training(Och, 2003) provided in MOSES tool kit itself. We used the model which was trained on the data with the end-of-sentence markers removed for declarative sentences. The Bleu score observed was 19.98. All these experiments were done on the language model trained on the training model. More experiments have to be conducted on the larger monolingual corpora. Table 3 shows the results of the experiments which involve the pre-processing steps.

## 7 Observations

In the course of our expriments, the following observations were made. The Bleu score did not show any significant improvement when the parser was used to reorder the sentences on the sources side. This falsifies our initial hypothesis that reordering relations learnt from the parse would be more generic and hence would improve the performance. The other pre-processing techniques like using a bilingual dictionary did not bring about any change in the Bleu score. One interesting observation was that upon deleting the end of sentence markers and storing them elsewhere the Bleu score showed a asignificant rise from 17.70 to 18.09. This is a considerable improvement when we consider the naivete of this pre-processing step. An improvement of 12.8% on the initial baseline was observed after the baseline model was tuned on the development set and tested on the test set.

## 8 Future Directions

Based on the above initial experiments we believe that reordering of the target language phrases improve substantially by tapping the available resources for English. These experiments are only a first step in improving the Bleu Score and much more has to be achieved.